\documentclass[10pt,twocolumn,letterpaper]{article}

\usepackage{cvpr}
\usepackage{times}
\usepackage{epsfig}
\usepackage{graphicx}
\usepackage{amsmath}
\usepackage{amssymb}
\usepackage{subcaption}
\usepackage{textcomp}

\usepackage[pagebackref=true,breaklinks=true,letterpaper=true,colorlinks,bookmarks=false]{hyperref}

\cvprfinalcopy 

\def\cvprPaperID{****} 

\ifcvprfinal\fi
\begin{document}

\title{View Classification and Object Detection in Cardiac Ultrasound to Localize Valves via Deep Learning}

\author{Derya Gol Gungor, Bimba Rao, Cynthia Wolverton, Ismayil Guracar\\
Siemens Healthineers\\
685 E Middlefield Rd,
Mountain View, CA 94043\\
{\tt\small \{derya.gol, bimba.rao, cynthia.wolverton, ismayil.guracar\}@siemens-healthineers.com}
}

\maketitle

\begin{abstract}
Echocardiography provides an important tool for clinicians to observe the function of the heart in real time, at low cost, and without harmful radiation.  Automated localization and classification of heart valves enables automatic extraction of quantities associated with heart mechanical function and related blood flow measurements. We propose a machine learning pipeline that uses deep neural networks for separate classification and localization steps. As the first step in the pipeline, we apply view classification to echocardiograms with ten unique anatomic views of the heart. In the second step, we apply deep learning-based object detection to both localize and identify the valves. Image segmentation based object detection in echocardiography has been shown in many earlier studies but, to the best of our knowledge, this is the first study that predicts the bounding boxes around the valves along with classification from 2D ultrasound images with the help of deep neural networks. Our object detection experiments applied to the Apical views suggest that it is possible to localize and identify multiple valves precisely. 
\end{abstract}

\section{Introduction}

Echocardiography (diagnostic cardiac ultrasound imaging) is routinely used to visualize the chambers and valves of the heart. Typically, it is combined with Doppler ultrasound to evaluate blood flow through valves and within chambers.  High frequency sound waves are transmitted into the body, and the received echoes from tissue are processed to produce both 2D images, and blood flow velocity estimates.  

This study focuses on heart valves which are critical components of the heart, namely the valves that control circulation between the chambers and aorta: Tricuspid Valve (TV), Mitral Valve (MV) and Aortic Valve (AV). TV is located in the right side of the heart, and MV is in the left side. These valves are called atrioventricular valves, since they connect the atria to the ventricles. In an ideal heart, blood flows through the both valves during diastole with contraction of the corresponding atrium and both close during systole with contraction of the corresponding ventricle to prevent regurgitation of blood from the ventricle to the atria. On  the other hand, the AV is responsible for controlling the blood circulation between the left ventricle and aorta, which is the main artery supplying oxygenated blood to the circulatory system. 

In the case of a heart with pathology,  the blood may flow backwards through the valve  if the valve does not close completely (regurgitation or insufficiency of the valve). MV and AV regurgitation affect more than 200.000 people per year in United States. 
Another significant valve abnormality is stenosis where the valve flaps become stiff resulting in narrowed valve openings and reduced blood flow. Similarly, Tricuspid atresia may limit blood flow because the valve is not formed properly, and a solid sheet of tissue blocks the passage between the chambers. Any significant valve insufficiency and abnormality can affect the quality of daily life, and may require significant treatment procedures. Untreated pathologies can result in enlargement of heart, heart rhythm problems (arrhythmia), heart failure or even death.  

Classification and localization of anatomy are key enabling technologies that open up doors to many solutions for Ultrasound techs as well as clinicians. Training and placement guidance for new and/or inexperienced users is an application that would tremendously benefit from these technologies. Successful localization and identification of valves would allow automatic highlighting and enhancing these organs in the image to make accurate measurements, guide procedures, place devices, etc.

\subsection{Related Work}

Deep neural networks are beginning to assist Echocardiagram image analysis. Automatic view classification is a popular application area since it is a basis for many other applications. Machine learning and image processing techniques for ultrasound image classification have been explored by many papers. Here, we focus only on those papers using deep neural networks. 
In \cite{gao2017fused}, two convolutional neural networks (CNNs) are combined to classify eight different views, namely Apical 2, 3, 4, 5 chamber, parasternal long axis (PLAX), parasternal short axis at aortic valve (PSAX-AoV), PSAX of papillary (PSAX-LV) and PSAX at mitral valve (PSAX-MV). In addition to brightness mode (B-mode) images, the temporal acceleration images are processed by a different network and the results are fused to obtain a final decision. This combination provided an average of 92\% accuracy for all views, with the lowest accuracy (71.4\%) in Apical 5 (mostly mixed with Apical 3 class) . Another study on view classification was done by ~\cite{Madani2018}. In addition to the classes listed above, they included PLAX RV-inflow, subcostal four-chamber, subcostal inferior vena-cava, subcostal aorta, suprasternal aorta, pulse-wave Doppler (PW), continuous-wave Doppler (CW) and motion mode (M-mode). Using the VGG-16 network\cite{VGGPaper}, they achieved 97.8\% overall accuracy in 15 different views with the accuracy for PLAX RV-inflow, specifically 86\% in video and 72\% on still images.This exceeds the prediction accuracy of a board-certified echo cardiographer. A more extended classification study was performed by~\cite{Zhang2018}, which also included subclasses of certain views. For example, in addition to the typical Apical 2 Chamber (A2C) view, the study included A2C plus occluded left atrium, and A2C plus occluded left ventricle. Due to the high correlations between classes, their average accuracy was 84\% on 23 views. However, if the results are considered in terms of broad classes such as PLAX, the accuracy they achieved was around 96\%. The classification network used was the 15 layer VGG network \cite{VGGPaper}. In addition to the view classification, they used deep networks also for image segmentation of cardiac chambers and also disease classification. However, image segmentation and disease detection are outside the scope of this paper. 

In the field of machine learning, object detection refers to the obtaining of bounding box coordinates for sub-images to identify and  localize multiple objects in a single image. This differs from image segmentation which performs pixel-wise classification to identify regions in an image. A detailed review of deep learning based object detection methods is given in \cite{zhao2018object}. From a high level perspective, there are two main approaches to object detection: region proposal based and regression/classification based. The first stage of the well-known region-based CNN technique (R-CNN)  \cite{girshick2014rich} is a region proposal generation technique. As an efficient alternative to exhaustive search, R-CNN uses a selective search algorithm \cite{uijlings2013selective} which iteratively merges small regions by hierarchical grouping according to their color spaces and similarity metrics. Then these regions are fed into a CNN for feature extraction. The features are fed into multiple SVM classifiers to provide class probabilities and also to a linear regressor to optimize the bounding box coordinates. Running multiple CNNs for each region proposal is computationally expensive. Therefore in Fast R-CNN \cite{girshick2015fast}, the order of image processing components is altered such that the feature extraction CNN is executed first, followed by  the region proposal network (RPN). The implication is that the CNN runs only one time over the entire input image to generate a feature vector. The output of CNN is connected to two fully-connected (FC) layers: one to produce bounding box coordinates (as regressor) and the other to produce object-ness probabilities (as classifier). Also the maximum number of regions is fixed a priori. In Faster RCNN \cite{ren2015faster}, a pre-trained region proposal network (RPN) is used to avoid the expensive selective search algorithm. Faster RCNN provides a more accurate and efficient architecture. However, some region-specific components still need to be applied hundreds of times per region proposal \cite{huang2017speed}. This is handled by R-FCN \cite{dai2016r}, where region crops are calculated on the final layer of the network. This  provides a significant  speed-up (2.5-20 times \cite{dai2016r} in tests with respect to Faster R-CNN). On the other hand, regression/classification based techniques such as Single-shot Detector (SSD) \cite{SSD} and YOLO \cite{YOLO} predict the bounding boxes and class probabilities all at once instead of a two step mechanism and thus may be more suitable for real-time applications. 

Several algebraic, signal processing and machine learning techniques have been proposed for tracking cardiac valves. In \cite{dukler2018automatic}, a non-negative matrix approximation approach has been used to detect and track the mitral valve in Apical 4 chamber views. This algorithmic approach does not require any labeling of the data but is limited to detecting the mitral valve only. Reference \cite{voigt2015robust} uses a machine learning approach for real-time tracking of mitral valve in 3D images for inteventional guidance. Their technique relies on the box estimator, based on \textit{marginal space learning (MSL) approach} \cite{zheng2009constrained} that predicts the presence of the MV location, orientation and scale in 3D images. MSL is three-step detector that applies probabilistic boosted-tree based classifiers multiple times to estimate different parameters. None of these studies uses deep neural networks. 

Convolutional networks in conjunction with object detection are commonly used in medical imaging for localization and segmentation of anatomical structures and organs. For example, \cite{de2017convnet} uses a specialized convolutional network called BoBNet (a variation of VGG network\cite{VGGPaper}) to predict bounding boxes around organs such as liver, heart, aorta applied to 3D computerized tomography images (CT) images. On the other hand, there is a lot of effort in fetal ultrasound imaging to identify the imaging plane and detect the structures being imaged. For example, in \cite{huang2017temporal} and \cite{sundaresan2017automated}, CNN based techniques are proposed to automatically localize a fetal heart. Another CNN based technique \cite{baumgartner2017sononet} is able to detect fetal standard planes and localize structures such as brain, spine, kidneys, lips, femur, etc.  from 2D ultrasound images.

In this paper, we illustrate how deep neural networks developed for the object detection problem perform on cardiac ultrasound images to detect the valves in different cardiac views. In the first section, we explain the image preprocessing and view classification applied as the initial step. Then, in the second section, we concentrate on the annotated data specific to object detection and mention the selected network for training. In the final section, we show the results of our experiments. 

\section{Materials and Methods}

\subsection{View Classification}
\begin{figure}[h]
\centering
\includegraphics[width=\linewidth]{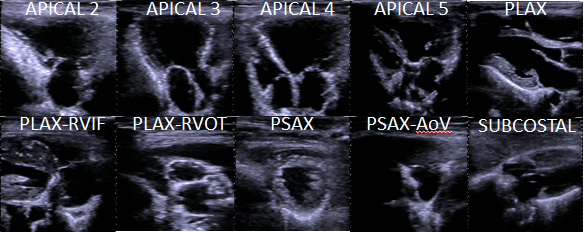}
\caption{Example images for each class used in view classification}
\label{fig:classes}
\end{figure}

View Classification of selected echocardiogram views is the first stage in our proposed machine learning pipeline. Views classified are as follows: Apical 2, Apical 3, Apical 4, Apical 5, Parasternal-long-axis (PLAX), PLAX-RVinflow (PLAX-RVIF), PLAX-RVoutflow (PLAX-RVOT), Parasternal-short-axis (PSAX)\footnote{PSAX view can be from Left Ventricle level or Mitral Valve. We observed that those two views look very similar and object detection to isolate the valve would produce similar results, thus we merged these two sub-classes.}, PSAX at the aortic valve level (PSAX-AoV), Subcostal of four-chamber, and Noise. The noise images for training are created by capturing images where the ultrasound probe is in contact with air, or in contact with ultrasound coupling gel only. An example image for each cardiac image class is shown in Fig.~\ref{fig:classes}.

\subsubsection{Annotateted Data}
The complete dataset includes 11,150 B-mode clips of 1-5 heartbeats from Acuson SC2000, Siemens Cardiac Ultrasound system (Mountain View, CA, USA).  All the clips are anonymized to remove any patient specific information. To ensure that the view classification testing is valid, care was taken to ensure that B-Mode clips from the same patient do not appear in both the training and the test data.  Because the data is anonymous, we have used acquisition date and time that was contained in the DICOM header files. We numbered the clips with a separate patient ID if there is 30 minutes gap between two closest acquisitions. This method may result in identifying two separate patients as the same person (if two studies were done back-to-back). However, it has low probability of splitting the images from a single patient into two parts.  With this method, we identified the total number of subjects to be at least 525. We needed this separation also to make sure that we do not include same patient's data in both training and validation/test to eliminate the bias. We partitioned 60\% of patients for training, 20\% for validation and 20\% for testing. 

\subsubsection{Preprocessing}
\begin{figure}
	\centering
	\begin{subfigure}[t]{0.25\textwidth}
		\centering
		\includegraphics[height=1.2in]{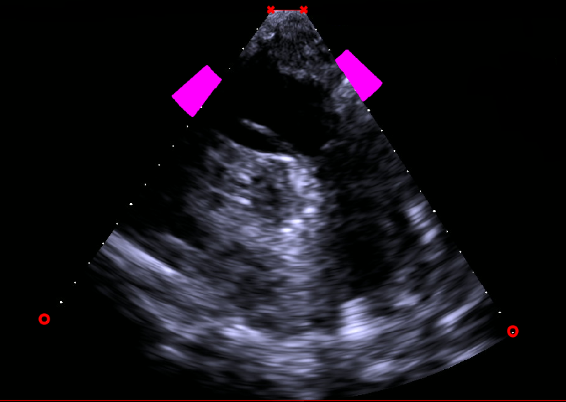}
		\caption{Before}
		\label{fig:scan_converted}
	\end{subfigure}
	\begin{subfigure}[t]{0.18\textwidth}
		\centering
		\includegraphics[height=1.2in]{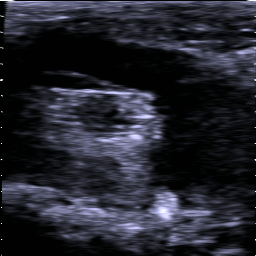}
		\caption{After}
		\label{fig:acoustic_converted}
	\end{subfigure}
	\caption{An example input image (a) before and (b) after pre-processing}
\end{figure}

The images were pre-processed in preparation for training. First, we divided the clips into frames (frame rate was 50-70 frames per second) and randomly selected 10 frames per heartbeat up to a maximum of 30 frames per clip. This allowed us to have 126,731 images for training, 38,148 for validation and 35,932 for testing. In typical ultrasound images, anatomical structures are shown in a polar coordinate system within a trapezoid-shaped area as shown in Fig.~\ref{fig:scan_converted}. There also exists some text related to system, acquisition or patient related information on the left and right side of rectangular images. In order to provide only necessary information to our network, the data is converted  from the display grid (the trapezoid shape) to a Cartesian grid in the first pre-processing step. To do this, the image scan depth and the trapezoid angles as shown in Fig.~\ref{fig:scan_converted} are automatically identified in MATLAB\textregistered (Natick, MA, USA). The red dots illustrate the extracted corners of the image and the magenta regions show the area scanned to find the angle of the trapezoid. Then the image within the trapezoid is transformed into Cartesian coordinate system via linear interpolation to give us a converted image as shown in Fig.~\ref{fig:acoustic_converted}. The images were resized to $256 \times 256$. The mean calculated from the training set images is extracted from each image. The grayscale images (maximum value of 255) were normalized to have values between 0-1 before feeding into the network. 

\begin{figure*}[h]
\centering
\includegraphics[width=\textwidth]{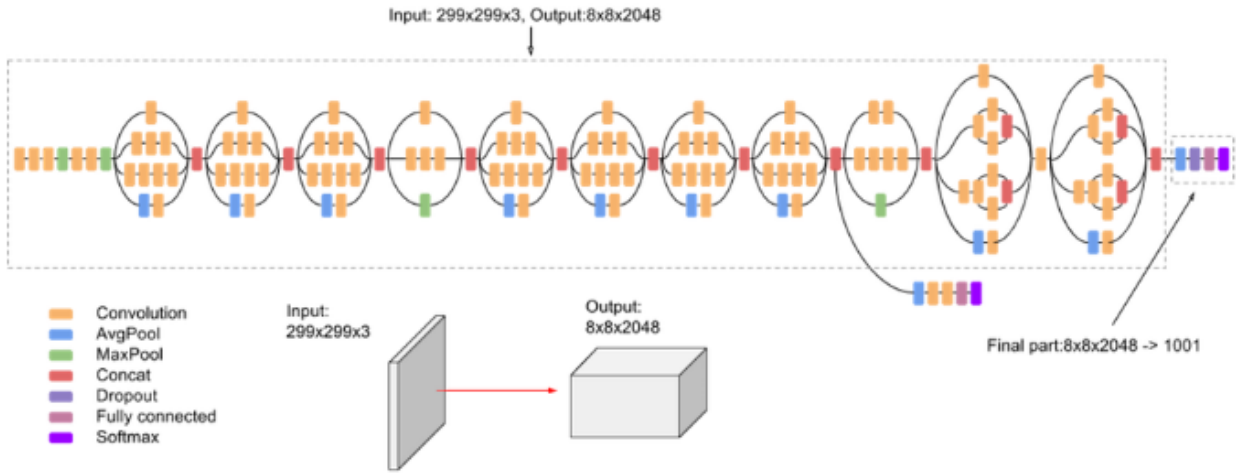}
\caption{InceptionV3 detailed diagram \protect\footnotemark}
\label{fig:inceptionV3}
\end{figure*}
\footnotetext{\url{https://cloud.google.com/tpu/docs/inception-v3-advanced}}

\subsubsection{Network}
We have adopted the \textbf{InceptionV3} network as described in \cite{szegedy2016rethinking}. The structure of InceptionV3 is shown in Figure~\ref{fig:inceptionV3}. Note that each convolutional layer follows batch normalization and activation units. Naive Inception modules include combinations of $1\times 1$, $3\times 3$, $5\times 5$ convolutional layers plus a $3\times 3$ pooling layer. The output of all those layers are then concatenated to create the input to the next layer. 
In InceptionV3, several factorization and dimensionality reduction techniques are used in Inception layers to increase computational efficiency. For example, most convolutional layers are preceded by a  $1\times 1$ block to reduce dimension along depth. Also, instead of using expensive large size convolutions (e.g. $5\times 5$ or $7\times 7$), multiple cascaded small-size (e.g. $3\times 3 $) are used to greatly reduce the number of parameters without loss of expressiveness \cite{szegedy2016rethinking}. Another technique to reduce computation is to use asymmetric factorization of convolutional layers that takes advantage of the separable property of convolution. For example, $3\times 3$ can be factorized into $3\times 1$ and $1\times 3$ cascaded layers. The resulting two layer network can provide 33\% computational efficiency improvement.

InceptionV3 starts with three cascaded convolutional layers, whose dimensions are $3\times 3\times 32$ (stride 2),  $3\times 3\times 32$ (stride 1), $3\times 3\times 64$ (stride 1), respectively. 
 It is followed with a $3\times 3$ (stride 2) maximum pooling layer. Then, two  cascaded convolutional layers again with filter sizes $1\times 1\times 80$ (stride 1) and $3\times 3\times 192$ (stride 1); this again is followed with a maximum pooling layer with $3\times 3$ (stride 2). After this, 15 different Inception modules that use all the computational techniques explained above follow. For an input size $299\times 299 \times 3$, the output size at the end of the complete network becomes $8\times 8 \times 2048$. 

In our experiments, we have used Python 3.6.5 from Anaconda (Austin, TX, USA), Keras 2.2.2, Keras Applications 1.0.5 and Keras Preprocessing 1.0.3 packages. Keras has been set to work in Tensorflow \cite{abadi2016tensorflow} backend. We have used tf-nightly-gpu (version 1.10.0) for Tensorflow (Google Inc, Mountain View, CA, USA). Training was run on a 64 bit Windows 10 system with Intel\textregistered
 Xeon\textregistered CPU E5-2640 processor, a single Nvidia\textregistered Geforce Titan X (12GB) GPU card and 32GB RAM. The training took around 24 hours for 20 epochs with a training batch size 64 ($\approx 40,000$ iterations). After each epoch, prediction on a validation set was performed and the model with lowest validation loss among 20 iterations was used to determine the final model. The loss function was the categorical cross-entropy function, used along with the Adadelta optimizer (learning rate=1, rho=0.95, epsilon=1e-8, decay=0). The class with maximum score was used as the final prediction value. 
We also applied random data augmentation with zoom range up to 15\%, shear range up to 3\%, height and width shift ranges up to 15\% , rotation angles up to 10 degrees, contrast range from -100 to 40. Please see the Results section for the classification results. 

\subsection{Object Detection}
In order to localize the heart valves, we have applied object detection training for Apical 2, Apical 3 and Apical 4 classes separately\footnote{The annotations for the other views are not yet complete.}. We have used Tensorflow's Object Detection API \cite{huang2017speed}, which includes implementation of well-known deep learning based networks (SSD, RFCN, Faster-RCNN) and the tools for easy training and testing. The system we used for training was the same system and same packages/software mentioned in the View Classification section. 

The number of classes for identification within the object detection portion of our pipeline depends on the view classification result obtained in in the first stage of the pipeline.  In the Apical 2 view, we have only the Mitral valve (MV) we are trying to correctly identify.  In Apical 3, we have MV and aortic valve (AV); and in Apical 4 we have MV and tricuspid valve (TV); and in Apical 5  we have left ventricle outflow tract (LVOT) just above AV. Thus, we train object detection networks separately for each cardiac view. 

\subsubsection{Annotated data}
As a basis for object detection, the B-Mode DICOM clips were annotated. Only one frame within a B-Mode DICOM clip was annotated, however we used the derived bounding box as ground truth for neighboring frames in the clip as well. Annotations were done on the frame in the heart cycle where the valve is completely closed. The annotation specifies three coordinate locations within the image: the center point where valve flaps touch when fully closed, and the left and right points where the valve connects to heart wall tissue. Since the annotation did not include top and bottom coordinates to define a ground truth bounding box, we selected a fixed height for all bounding boxes which is large enough to encompass the valve when it is fully open. In addition, the width of the valves in annotations can have variation since there is no specific line to distinguish the valves from the connecting tissue. Precise annotations on each frame in the clip could have improved accuracy.  

\subsubsection{Network}
\begin{figure}
\centering
\includegraphics[width=0.90\linewidth]{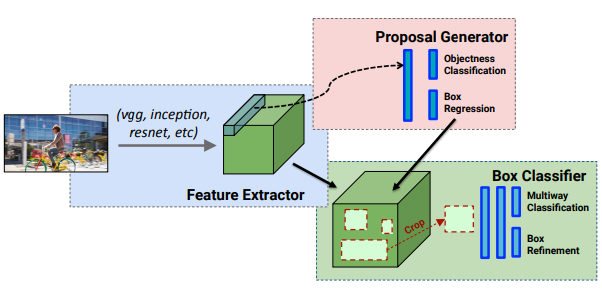}
\caption{Diagram of R-CNN (image from \cite{huang2017speed})}
\label{fig:rcnn}
\end{figure}

The meta-structure, we adopted was Faster R-CNN \cite{ren2015faster}, with a high-level diagram shown in Fig.\ref{fig:rcnn}. R-CNN networks consists of two stages. The first stage is called \textit{region proposal network (RPN)}, where the features are extracted from the intermediate layers of the networks such as Inception, ResNet or VGG. These features are given to a \textit{region proposal generator} that outputs the bounding box coordinates and object-ness scores of fixed number of regions (e.g.300). In the second stage, a cropped set of sub-images created using the region proposals is fed to the remainder of the feature extractor network to output the predicted class and refined bounding box coordinates. Since this operation is done separately for each proposed region, the speed of the Faster R-CNN is highly dependent on the number of region proposals selected. Details about the loss functions and speed/accuracy comparisons of different feature extractors and meta-structures can be found in \cite{huang2017speed}. 

\begin{figure*}
	\centering
	\begin{subfigure}[t]{0.49\textwidth}
		\centering
		\includegraphics[width=0.95\linewidth]{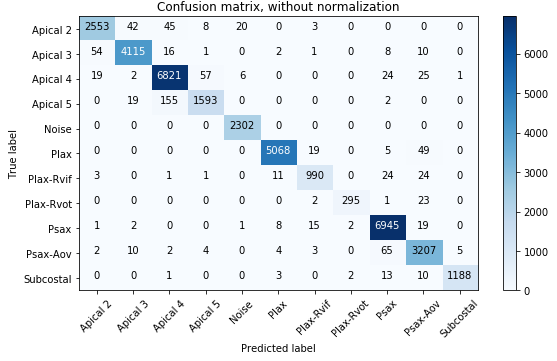}
		\label{fig:confusion_matrix}
	\end{subfigure}
	\begin{subfigure}[t]{0.49\textwidth}
		\centering
		\includegraphics[width=0.95\linewidth]{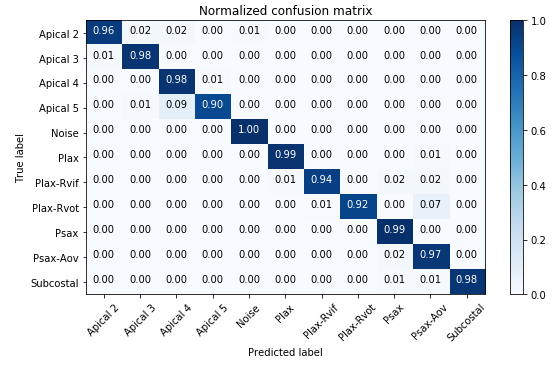}
		\label{fig:confusion_matrix_norm}
	\end{subfigure}
	\caption{Confusion matrix for test set: (a) distribution in frames (b) normalized by the number images per class}
	\label{fig:CM}
\end{figure*}

We specifically used \textbf{Faster-RCNN with ResNet101} \cite{he2016deep}. This network has been shown to be slower than other deep learning based networks such as SSD and R-FCN but provides more accurate results \cite{huang2017speed} based on experiments done on the Microsoft COCO dataset \cite{COCO}. The speed and accuracy rates of this network are highly dependent on parameters such as input image size and the number of bounding box proposals. In our applications we used low-resolution images ($256\times256$) but selected the number of proposals as 300. 

We started with a pre-trained faster-RCNN network on the COCO dataset (the checkpoint was downloaded from the Model Zoo website\footnote{\url{https://github.com/tensorflow/models/blob/master/research/object_detection/g3doc/detection_model_zoo.md}}). The data augmentation options we have used were rgb-to-gray, random-horizontal flip, random-adjust-brightness, random-adjust-contrast, random-crop-and-pad-image (min-area: 0.5, min-padded-size-ratio: [1,1], max-padded-size-ratio: [2,1]). Because random-crop-and-pad image augmentation option may result in a change of image size, we have provided batch size equal to 1 in training. Other parameters are kept unchanged from the values provided in the configuration file that comes with the pre-trained model but we also provided them in the supplementary material. The same images provided to the classification network were used as input to the object detection network. The results of the experiments are provided in the next section. 

\section{Results}

\subsection{View Classification}
\begin{figure}
\centering 
\includegraphics[width=0.80\linewidth]{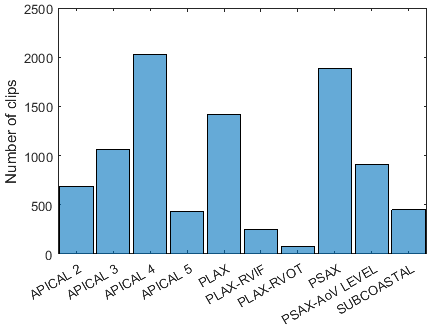}
\caption{Distribution of dicom clips in overall data}
\label{fig:distribution}
\end{figure}

The distribution of DICOM clips in our training, validation and test data belonging to each class are shown in Fig.~\ref{fig:distribution}. As can be seen there is a significant imbalance between classes. The exact numbers are 686 (Apical 2), 1061 (Apical 3), 2028 (Apical 4), 428 (Apical 5), 1416 (PLAX), 250 (PLAX-RVIF),  74 (PLAX-RVOT), 1882 (PSAX), 909 (PSAX-AoV), 451 (Subcostal). 
The overall accuracy we obtained was 97.62\% on the test set. The confusion matrix obtained from the test set are shown in Fig.~\ref{fig:CM}. The diagonal elements show the number of correct predictions whereas off-diagonal elements show the number of mis-classified images. The lowest accuracy we obtained was in the Apical 5 class, mostly because of high correlation to Apical 4 images. In addition, while the heart is contracting, the chamber appearing in the center (the aorta) can become very small in some frames, which makes the image look like an Apical 4 view. Also, we observed that a zoomed Apical 5 may look more like Apical 3. Next lowest accuracy values were obtained in the PLAX-RVIF and PLAX-RVOT views. Images in these classes exhibit large variations from patient to patient. Additional training data representing all these variations was yet not available. 

\begin{table*}[ht]
\caption{Evaluation results of object detection experiments} 
\centering 
\begin{tabular}{c c c c c c} 
\hline\hline 
Class &  \# test images &mAP (IoU:0.50:0.95) &  mAP (IoU:0.50)  & mAP (IoU:0.75) & mAR (IoU:0.50:0.95) \\ [0.5ex] 
\hline 
Apical 2 & 2164 & 0.151 & 0.493 & 0.041 & 0.451 \\ 
Apical 3 & 2819 & 0.170 & 0.547 & 0.042 & 0.450 \\
Apical 4 & 5303 & 0.343 & 0.896 & 0.146 & 0.528  \\ [1ex] 
\hline 
\end{tabular}
\label{table:obj_det_results} 
\end{table*}

\begin{figure*}[h]
\centering 
\includegraphics[width=\linewidth]{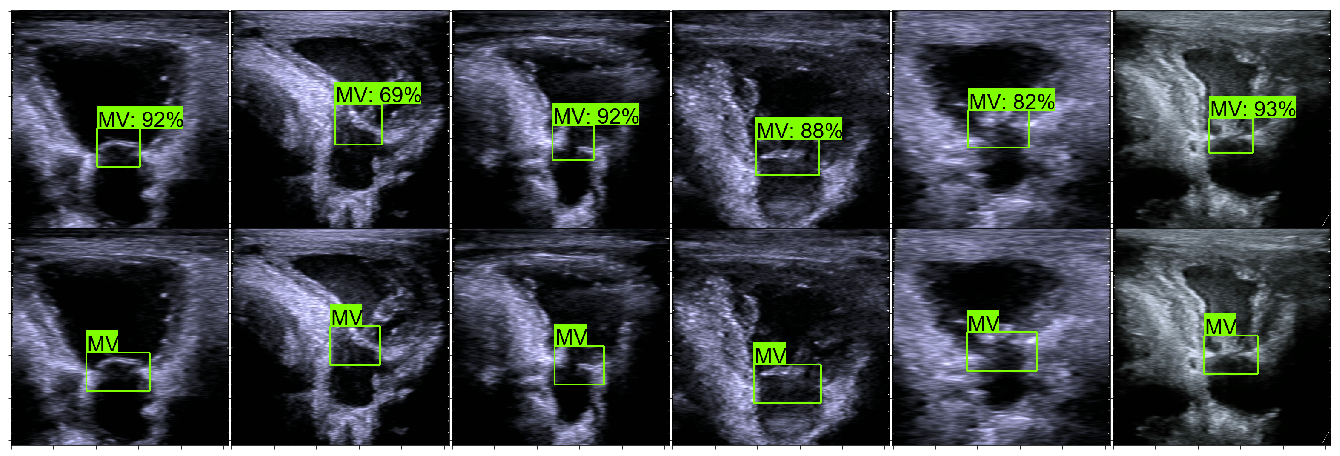}
\caption{Predicted bounding boxes (top row) and ground truths (bottom row) for Apical 2 view test images. Green represents Mitral valve (MV) with prediction scores in percentage.}
\label{fig:res_a2c1}
\end{figure*}

\begin{figure*}[h]
\centering 
\includegraphics[width=\linewidth]{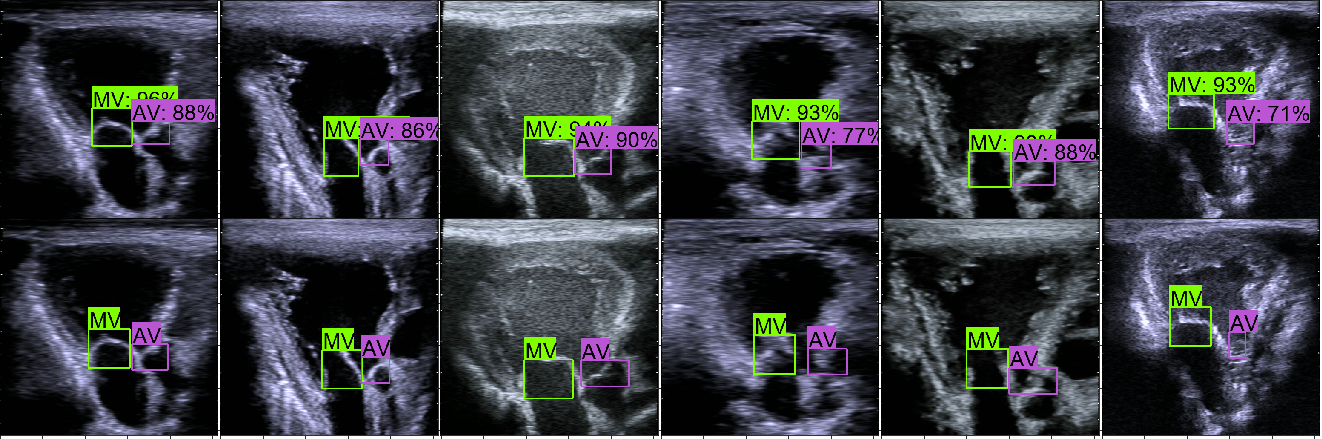}
\caption{Predicted bounding boxes (top row) and ground truths (bottom row) for Apical 3 view test images. Green represents Mitral valve (MV) and cyan represents Aortic Valve (AV) with prediction scores in percentage.}
\label{fig:res_a3c1}
\end{figure*}

\subsection{Object Detection}
In the evaluation of the object detection step, we have used mean average precision (mAP) and mean average recall (mAR) values calculated for the given intersection-over-union ratios (IoU). The evaluation results are given in Table \ref{table:obj_det_results}. Here, mAP (IoU:0.50:0.95) corresponds to average mAP calculated over a range of IoU = 0.50:0.05:0.95. This metric is MS COCO's standard detection metric. On the other hand, mAP (IoU:0.50) corresponds to mAP calculated at IoU=0.50 (PASCAL VOC's metric \cite{pascal-voc-2012}). We have also shown results for mAP (IoU:0.75)  and  mAR (IoU:0.50:0.95). As can be seen from Fig.~\ref{table:obj_det_results}, the best results are obtained for Apical 4 views because there are two times more samples available in training, and also MV and TV are larger valves than AV, and therefore easier to detect. On the other hand, the Apical 2 view has the largest MV but we have obtained low mAP values. We believe that this is simply because of the variations in ground truths mentioned above and this is more pronounced in Apical 2, where the valve appears big. 

We also show some visual examples of bounding box detection applied to six different images selected from the test set per each cardiac view. These images were hand-picked to represent different B-mode dynamic range, shifts, zoom factors, rotation and noise levels. In all of the figures corresponding to different views, the top row represents the detection results of maximum score ($>0.5$) and the bottom row represents the ground truth. We used the same color scheme (green) for MV as it appears in all of the Apical views we examined. AV appearing in Apical 3 view is shown in purple, and TV in Apical 4 view shown in cyan. 

Results for Apical 2 view are illustrated in Fig.~\ref{fig:res_a2c1}.  As can be seen, the valves were detected precisely in all the test images. The lowest score was 69\% obtained in the second test image since there is more structure appearing around the valve, probably due to the imaging plane being close the wall of the heart. 

Results for Apical 3 view are shown in Fig.~\ref{fig:res_a3c1}. The scores for the aortic valve are overall lower than the scores of mitral valve because it is a smaller valve and sometimes hardly visible in especially noisy images such as in the last column of Fig.~\ref{fig:res_a3c1}. 

\begin{figure*}[h]
\centering 
\includegraphics[width=\linewidth]{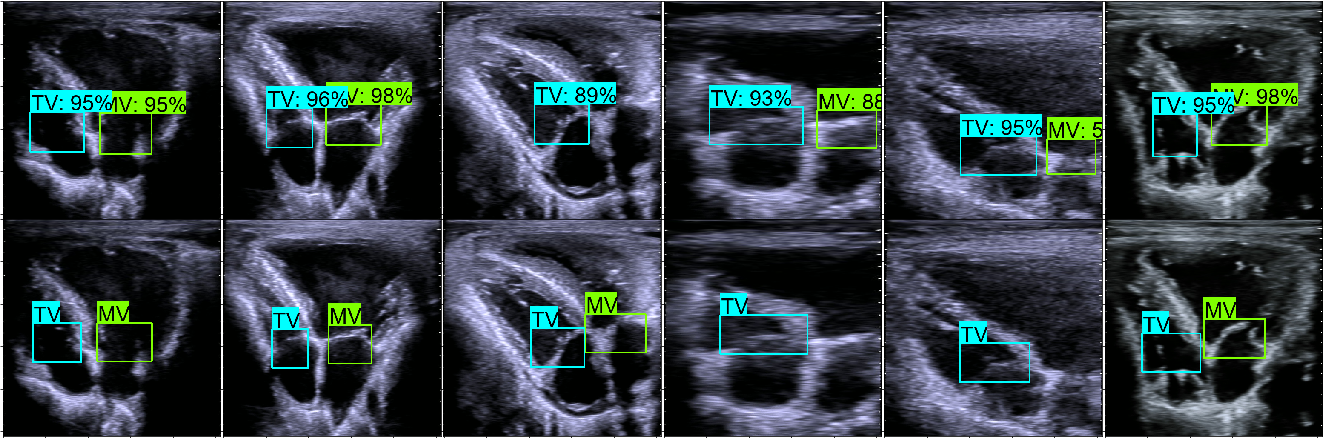}
\caption{Predicted bounding boxes (top row) and ground truths (bottom row) for Apical 4 view test images. Green represents Mitral valve (MV) and cyan represents Tricuspid Valve (TV) with prediction scores in percentage.}
\label{fig:res_a4c1}
\end{figure*}

Results for Apical 4 view are shown in Fig.~\ref{fig:res_a4c1}. Our annotations did not cover the valves appearing only partially (e.g. images in forth and fifth column of  Fig.~\ref{fig:res_a4c1}). However, the network was able to detect these valves as there is crop-and-pad option in our data augmentation, that may create such examples of partial valve images in the training data. When the heart is on a rotated plane, the MV appears smaller in size, which also decreases its detection probability with very low scores as seen in the third and fifth columns of Fig.~\ref{fig:res_a4c1}.

\section{Conclusion}
We have presented an end-to-end deep learning based pipeline that includes classification and object detection modules for the localization and identification of valves in Apical views of cardiac ultrasound images.  To the best of our knowledge, this is the first paper that uses state-of the art deep learning based object detection networks for this specific application whereas previous studies use deep networks for segmentation. Our results suggest that it is possible to accurately locate and classify the valves using object detection techniques. For future work, we plan to add more training data with more precise annotations to increase the accuracy in Apical views. We also plan to extend the view classification to cover more views (such as other subcostal views and suprasternal) and apply object detection to the other cardiac views such as PLAX, etc. 

{\small
\bibliographystyle{ieee}
\bibliography{aibib}
}

\end{document}